you

\documentclass[letterpaper, 10 pt, conference]{ieeeconf}  

\IEEEoverridecommandlockouts                              

\usepackage{graphics} 
\usepackage{times} 
\usepackage{amsmath} 
\usepackage{amssymb}  
\usepackage{times}
\usepackage{epsfig}
\usepackage{graphicx}
\usepackage{amsmath}
\usepackage{amssymb}
\usepackage{stfloats}
\usepackage{amssymb}
\usepackage[noend]{algpseudocode}
\usepackage{algorithm}
\usepackage{xspace}
\usepackage{color}
\usepackage{adjustbox}
\usepackage{array}
\usepackage{hyperref} 
\usepackage{amsfonts}
\usepackage{booktabs}
\usepackage{siunitx}
\usepackage{multirow}
\usepackage{colortbl}
\usepackage[table, svgnames, dvipsnames]{xcolor}

\DeclareMathAlphabet{\mathpzc}{OT1}{pzc}{m}{it}

\newcommand{\mathkomma}{\quad ,}
\newcommand{\mathpunkt}{\quad .}

\newcommand{\snx}[1]{\textit{SalsaNext }{#1}} 
\newcommand{\snxk}[1]{\textit{SalsaNext, }{#1}} 
\newcommand{\snxp}[1]{\textit{SalsaNext. }{#1}} 
\newcommand{\sn}[1]{\textit{SalsaNet }{#1}} 
\newcommand{\snk}[1]{\textit{SalsaNet, }{#1}} 
\newcommand{\ls}[1]{\textit{Lov\'{a}sz-Softmax }{#1}}

\newcommand{\sk}[1]{Semantic-KITTI {#1}}  
\newcommand\Tstrut{\rule{0pt}{2.6ex}}         
\newcommand\Bstrut{\rule[-0.9ex]{0pt}{0pt}}   

\newcolumntype{R}[2]{%
    >{\adjustbox{angle=#1,lap=\width-(#2)}\bgroup}%
    l%
    <{\egroup}%
}

\newcommand\scalemath[2]{\scalebox{#1}{\mbox{\ensuremath{\displaystyle #2}}}}

\DeclareMathOperator*{\argmin}{arg\,min}

\def\eg{e.g.\@\xspace}
\def\ie{i.e.\@\xspace}

\title{\LARGE \bf
SalsaNext: Fast, Uncertainty-aware Semantic Segmentation \\ of LiDAR Point Clouds for Autonomous Driving
}

\author{Tiago Cortinhal$^{1}$, George Tzelepis$^{2}$ and Eren Erdal Aksoy$^{1,2}$
\thanks{The research leading to these results has received funding from the Vinnova FFI project SHARPEN, under grant agreement no. 2018-05001.}
\thanks{$^{1}$Halmstad University, School of Information Technology, Center for Applied Intelligent Systems Research, Halmstad, Sweden.} 
\thanks{$^{2}$Volvo Technology AB, Volvo Group Trucks Technology, Vehicle Automation, Gothenburg, Sweden.
}%
}

\begin{document}

\maketitle
\thispagestyle{empty}
\pagestyle{empty}

\begin{abstract}

In this paper, we introduce   \snx   for the uncertainty-aware semantic  segmentation of a full 3D LiDAR point cloud in real-time.
\snx is the \textit{next} version of \sn ~\cite{salsanet2020} which has an encoder-decoder architecture where the encoder unit has a set of ResNet blocks and the decoder part combines upsampled features from the residual blocks. 
In contrast to   \snk we   introduce a new context module, replace the ResNet encoder blocks with   a new residual dilated convolution stack with   gradually increasing receptive fields and add the \textit{pixel-shuffle} layer in the decoder.
Additionally, we switch from stride convolution to average pooling and also apply central dropout treatment.
To directly optimize the Jaccard index, we further combine  the weighted cross entropy loss  with \ls loss \cite{berman2018lovasz}.
We finally inject a Bayesian treatment to compute  the \textit{epistemic}  and  \textit{aleatoric}   uncertainties for each point in the cloud.
%
We provide a thorough quantitative  evaluation on the \sk dataset \cite{semantickitti}, which demonstrates that the proposed \snx outperforms other state-of-the-art semantic segmentation networks and ranks first on the \sk leaderboard.
We also release our source code \href{https://github.com/TiagoCortinhal/SalsaNext}{https://github.com/TiagoCortinhal/SalsaNext}.


\end{abstract}

\section{Introduction}

Scene understanding is an essential prerequisite for  autonomous vehicles.
Semantic segmentation helps gaining a rich understanding of the scene by predicting a meaningful class label   for each individual sensory data point. 
Achieving such a fine-grained semantic prediction in real-time   accelerates reaching the full autonomy to a great extent.

Safety-critical systems, such as self-driving vehicles, however,  require not only   highly  accurate but also   reliable predictions with a consistent measure of uncertainty. 
This is because the   quantitative uncertainty measures can be propagated to the subsequent units,  such as decision making modules to lead to safe manoeuvre  planning or emergency braking,   which is of utmost importance in safety-critical systems.
Therefore, semantic segmentation   predictions integrated with reliable confidence estimates can significantly reinforce the concept of safe autonomy.

Advanced deep neural networks recently had a quantum jump in generating accurate and reliable semantic segmentation with real-time performance. Most of these approaches, however, rely on the camera images~\cite{kendall2015bayesian,FastSCNN2019}, whereas relatively fewer contributions have discussed the semantic segmentation of 3D LiDAR data~\cite{SqueezesegV01,rangenetpp}. The main reason is that unlike camera images  which provide dense measurements in a grid-like structure,   LiDAR point clouds are relatively sparse, unstructured, and have non-uniform sampling, although LiDAR scanners have a wider field of view and return more accurate distance measurements. 

\begin{figure}[!t]
    \centering
    \includegraphics[scale=0.49]{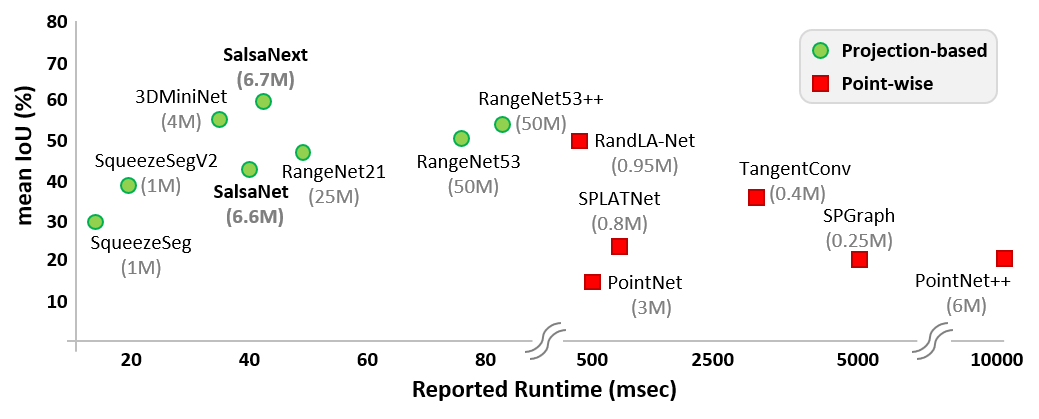}
    \caption{Mean IoU versus runtime plot for the state-of-the-art 3D point cloud semantic segmentation networks on the Semantic-KITTI dataset~\cite{semantickitti}. Inside parentheses are given the total number of network parameters in Millions. All deep networks visualized here use only 3D LiDAR point cloud data as input.   Note that only the published methods  are considered. }
    \label{fig:meanIoUvsRunTime}
\end{figure}

As comprehensively described in \cite{survey3DPC2019}, there exists two mainstream deep learning  approaches addressing the semantic segmentation of 3D LiDAR data only: point-wise and projection-based neural networks (see Fig.~\ref{fig:meanIoUvsRunTime}). The former approach operates directly on the raw 3D points   without requiring any pre-processing step,  whereas the latter projects the point cloud into various formats such as 2D image view or high-dimensional volumetric representation. As illustrated in Fig.~\ref{fig:meanIoUvsRunTime}, there is a clear split between these two approaches in terms of accuracy, runtime and memory consumption. 
For instance, projection-based approaches (shown in green circles in Fig.~\ref{fig:meanIoUvsRunTime}) achieve the state-of-the-art accuracy while running significantly faster. Although point-wise networks (red squares) have slightly lower number of parameters, they  cannot   efficiently scale up to large point sets due to the limited processing capacity, thus, they take a longer runtime. 
%
It is also highly important to note that  both  point-wise and projection-based approaches in the literature  lack  uncertainty measures, \ie confidence scores, for their predictions.

In this work, we introduce a novel neural network architecture to perform uncertainty-aware semantic  segmentation of a full 3D LiDAR point cloud in real-time.
Our proposed network is built upon the \sn model~\cite{salsanet2020}, hence, named   \snxp 
The \sn model has an encoder-decoder skeleton where the encoder unit consists of a series of ResNet blocks and the decoder part upsamples and fuses features extracted in the residual blocks. 
In the here proposed \snxk our contributions lie in the following aspects:
\begin{itemize}
\item To capture the global context information in the full $360^\circ$ LiDAR scan,  we introduce a new context module before encoder, which consists of a residual dilated convolution stack fusing receptive fields at various scales. 
\item To increase the receptive field, we replaced the ResNet block in the encoder with a novel combination of a  set of dilated convolutions (with a rate of 2) each of which has different kernel sizes ($3, 5, 7$). We further concatenated the convolution outputs and combined with   residual connections yielding a branch-like structure.
\item To avoid any checkerboard artifacts in the upsampling process, we replaced the transposed convolution layer in the \sn decoder with a \textit{pixel-shuffle} layer~\cite{pixelshuffle} which directly leverages on the feature maps to upsample the input with less computation.
\item To boost the roles of very basic  features (\eg edges and curves) in the segmentation process, the dropout treatment was altered by omitting the first and last network layers in the dropout process. 
\item To have a lighter model, average pooling was employed instead of having stride convolutions in the encoder.
\item To enhance the segmentation accuracy by optimizing the mean intersection-over-union score, \ie the Jaccard
index, the weighted cross entropy loss in \sn was combined with the \ls loss \cite{berman2018lovasz}.
\item To further estimate the \textit{epistemic} (model) and  \textit{aleatoric} (observation) uncertainties for each 3D LiDAR point, the deterministic \sn model was transformed into a stochastic  format by applying the Bayesian treatment.
\end{itemize}  

All these contributions form the here introduced \snx model which is the probabilistic derivation of the \sn with a significantly better segmentation performance. 
The input of \snx is the rasterized image of the full LiDAR scan, where each image channel stores position, depth, and intensity cues in the panoramic view format. The final network output is the point-wise classification scores together with uncertainty measures.

To the best of our knowledge, this is the first work showing the both   \textit{epistemic}   and  \textit{aleatoric} uncertainty estimation on the   LiDAR point cloud segmentation task. 
Computing both uncertainties is of utmost importance in safe autonomous driving since the \textit{epistemic} uncertainty can indicate the limitation of the segmentation model while the  \textit{aleatoric} one highlights the sensor observation noises for segmentation.
  
Quantitative and qualitative experiments on the \sk dataset \cite{semantickitti} show that the proposed \snx significantly outperforms other state-of-the-art     networks in terms of pixel-wise segmentation accuracy while having much fewer parameters, thus requiring less computation time. \snx ranks first place on the \sk leaderboard.

Note that we also release our source code and trained model to encourage further research on the subject.

\section{Related Work}

In this section, recent works in  semantic segmentation of 3D point cloud data will be summarized.
This will then be followed by  a brief review of the literature related to Bayesian neural networks for uncertainty estimation.
 
\subsection{Semantic Segmentation of 3D Point Clouds}
\label{sec:semseg}
 
Recently, great progress has been achieved  in semantic segmentation of 3D LiDAR point clouds  using deep neural networks \cite{salsanet2020,SqueezesegV01,rangenetpp,SqueezesegV02,PointSeg18}.
The core distinction between these advanced methods lies not only in the network design but also in the representation of the point cloud  data.

Fully convolutional networks \cite{fcn2016}, encoder-decoder structures \cite{Zhang2018}, and multi-branch models \cite{FastSCNN2019}, among others, are the mainstream network   architectures used for semantic segmentation. Each network type has a unique way of encoding features at various levels, which are then fused to recover the spatial information. 
Our proposed \snx follows the encoder-decoder design as it showed promising  performance in most state-of-the-art methods~\cite{SqueezesegV01,SqueezesegV02,Unet}. 

Regarding the representation of   unstructured and  unordered 3D LiDAR points,  there are two common approaches as depicted in Fig.~\ref{fig:meanIoUvsRunTime}: point-wise representation   and  projection-based rendering. We refer the interested readers to \cite{survey3DPC2019} for more details on the 3D data representation.

Point-wise methods  \cite{pointnet,pointnetpp} directly process the raw irregular 3D points without applying any additional transformation or pre-processing.
Shared  multi-layer perceptron-based PointNet~\cite{pointnet}, the subsequent work PointNet++~\cite{pointnetpp}, and \textit{superpoint} graph SPG networks~\cite{SPGraph} are considered in this group. 
Although such methods are powerful on small point clouds, their processing capacity and memory requirement, unfortunately, becomes inefficient when it comes to     the full $360^\circ$ LiDAR scans. 
To accelerate point-wise operations, additional cues, \eg from camera images, are employed as successfully introduced in \cite{Frustum2017}.

Projection-based methods instead transform the 3D point cloud into various formats such as  voxel cells \cite{Zhang2018,VoxelNet18,SEGCloud2017}, multi-view representation \cite{Felix17}, lattice structure \cite{SPLATNet,LatticeNet},  and rasterized images \cite{salsanet2020,SqueezesegV01,SqueezesegV02,SqueezesegV03}. 
In the multi-view representation, a 3D point cloud is projected onto multiple 2D surfaces from various virtual camera viewpoints. Each view is then processed by a multi-stream network as in \cite{Felix17}. 
In the lattice structure, the raw unorganized point cloud is interpolated to a permutohedral sparse
lattice where bilateral convolutions are applied to occupied lattice sectors only \cite{SPLATNet}. 
Methods relying on the voxel representation  discretize the 3D space into 3D volumetric space (\ie voxels) and assign each point to the corresponding voxel \cite{Zhang2018,VoxelNet18,SEGCloud2017}. Sparsity and irregularity in point clouds, however,  yield redundant computations in voxelized data since many voxel cells may stay empty. 
A common attempt to overcome the sparsity in LiDAR data is to project 3D point clouds into 2D image space either in the top-down Bird-Eye-View  \cite{salsanet2020,rt3d2018,ComplexYolo} or spherical Range-View (RV) (\ie panoramic view) 
 \cite{rangenetpp,SqueezesegV01,SqueezesegV02,SqueezesegV03,3Dmininet,PointSeg18} formats.
Unlike point-wise and other projection-based approaches, such 2D rendered image representations are more compact, dense and computationally cheaper as they can be processed by standard 2D convolutional layers. Therefore, our \snx model initially projects the LiDAR point cloud into 2D RV image   generated by mapping each 3D point onto a spherical surface.


Note that in this study we focus on semantic segmentation of LiDAR-only data and thus ignore multi-model approaches that fuse,  \eg LiDAR and camera data as   in \cite{Frustum2017}.

\subsection{Uncertainty Prediction with Bayesian Neural Networks}
Bayesian Neural Networks (BNNs) learn approximate distribution on the weights  to further generate uncertainty estimates, \ie prediction confidences.
There are two types of uncertainties: \textit{Aleatoric} which can quantify the intrinsic uncertainty coming from the observed data,  and \textit{epistemic} where the model uncertainty is estimated by inferring with the posterior weight distribution, usually through Monte Carlo sampling. 
Unlike \textit{aleatoric} uncertainty, which captures the irreducible noise in the data, \textit{epistemic} uncertainty can be reduced by gathering more training data.
For instance, segmenting out an object that has relatively fewer training samples in the dataset may lead to high \textit{epistemic} uncertainty, whereas high \textit{aleatoric} uncertainty may rather occur on segment boundaries or distant and occluded objects due to noisy sensor readings which are inherent in sensors.
Bayesian modelling helps estimating both uncertainty types.



Gal \textit{et al.}~\cite{gal2016dropout} proved that dropout can be used as a Bayesian approximation to estimate the uncertainty in classification, regression and reinforcement learning tasks while this idea was also extended to semantic segmentation of RGB images by Kendall \textit{et al.}~\cite{kendall2015bayesian}. 
Loquercio \textit{et al.}~\cite{segu2019general} proposed a framework which extends the dropout approach by propagating the uncertainty that is produced from the sensors through the activation functions without the need of retraining.
Recently, both uncertainty types were applied to 3D point cloud object detection \cite{feng2018towards} and   optical flow estimation \cite{ilg2018uncertainty} tasks. 
To the best of our knowledge, BNNs have not been employed in modeling the uncertainty of semantic segmentation of 3D LiDAR point clouds, which is one of the main contributions in this work.
 
In this context, the closest work to ours is \cite{Zhang2019} which introduces a probabilistic  embedding space for point cloud instance segmentation. This approach, however, captures neither the aleatoric nor the epistemic uncertainty but rather predicts the uncertainty between the point cloud embeddings.
Unlike our method, it has also not been shown how the aforementioned work can scale up to large and complex LiDAR point clouds. 

\section{Method}
In this section, we give a detailed description of our  method starting with the point cloud representation. We then continue with the network architecture, uncertainty estimation, loss function, and training details.

\subsection{LiDAR Point Cloud Representation}
\label{sec:projection}

As in \cite{rangenetpp}, we project the unstructed 3D LiDAR point cloud  onto a spherical surface to generate the LIDAR's native Range View (RV) image. This process leads to dense and compact point cloud representation which  allows standard convolution operations. 

In the 2D RV image, each raw LiDAR  point $(x,y,z)$ is mapped to an image coordinate  ($u,v$) as  
\[\begin{pmatrix}
\mathit{u}  \\
\mathit{v}
\end{pmatrix}
=
\begin{pmatrix}
\frac{1}{2} [1-\arctan(y,x)\pi^{-1}]\mathit{w}  \\
 [1-(\arcsin(z,r^{-1})+f_{down})f^{-1}]\mathit{h}
\end{pmatrix}
\mathkomma\]

where $\mathit{h}$ and $\mathit{w}$ denote the height and width of the projected image, $\mathit{r}$ represents the range of each point as  $r=\sqrt{x^2+y^2+z^2}$ and $\mathit{f}$ defines the sensor vertical field of view as $\mathit{f} = |f_{down}| + |f_{up}|$.   

Following the work of~\cite{rangenetpp}, we considered the full  $360^\circ$ field-of-view in the projection process. During the projection, 3D point  coordinates ($x,y,z$), the intensity value ($i$) and the range index ($r$)  are  stored as separate RV image channels. This   yields a $[\mathit{w} \times \mathit{h} \times 5]$ image to be fed to the network.

\begin{figure*}[!t]
    \centering
    \includegraphics[scale=0.6]{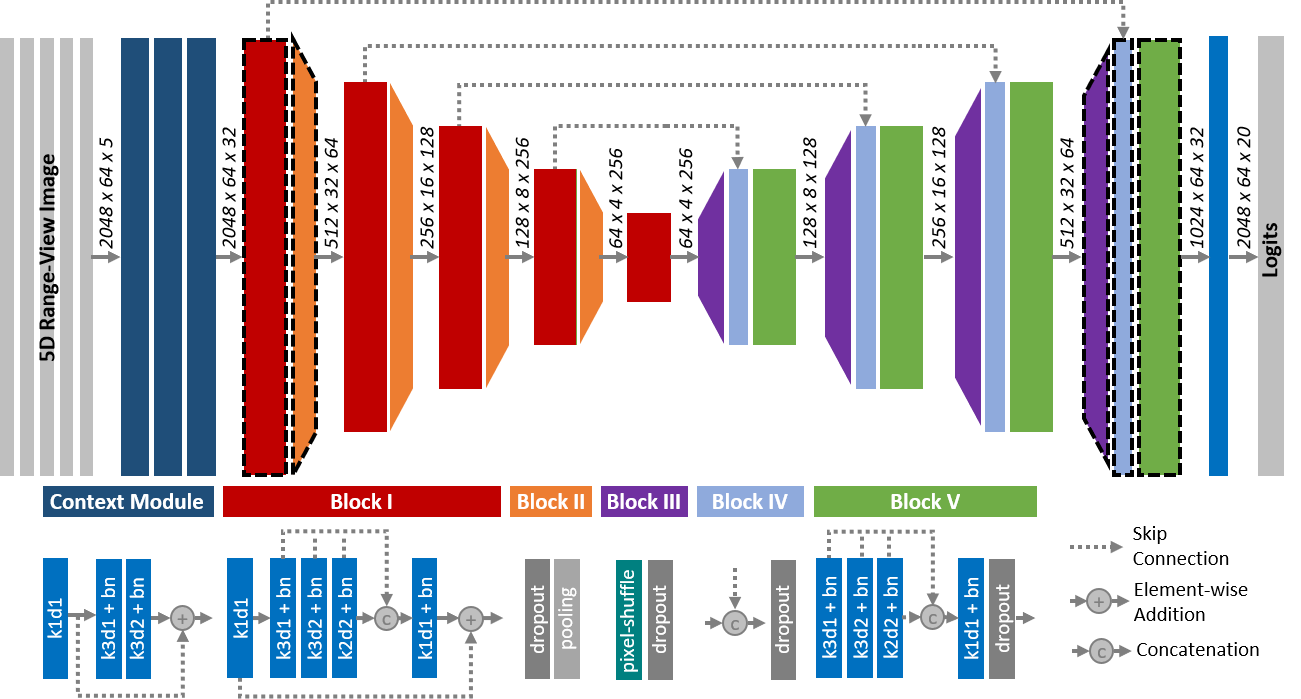}
    \caption{Architecture of the proposed \textit{SalsaNext} model.  Blocks with dashed edges indicate  those that do not employ the dropout. The layer elements $k, d,$ and $bn$ represent the kernel size, dilation rate and batch normalization, respectively.   
    }
    \label{fig:salsanext}
\end{figure*}

\subsection{Network Architecture}

The   architecture of the proposed \snx is illustrated in Fig.~\ref{fig:salsanext}. 
The input to the network is an RV image projection of the point cloud as described in section~\ref{sec:projection}.

\snx  is built upon the base  \sn model \cite{salsanet2020} which follows the standard encoder-decoder architecture with a bottleneck compression rate of 16. The original \sn encoder contains a series of ResNet blocks \cite{resnet2016} each of which 
is followed by dropout and downsampling layers. 
The decoder blocks apply  transpose convolutions and fuse upsampled features with that of the early residual blocks via skip connections. To further exploit descriptive spatial cues, a stack  of convolution is inserted  after the skip connection. 
As illustrated in Fig.~\ref{fig:salsanext},   we in this study  improve  the base structure of \sn with the following contributions:



\textbf{Contextual Module}: 
One of the main issues with the semantic segmentation is the lack of contextual information throughout the network.
The global context information gathered by larger receptive fields plays a crucial role in learning complex correlations between classes \cite{FastSCNN2019}.  
To aggregate the context information in different regions, we place a residual dilated convolution stack that fuses a larger receptive field with a smaller one by adding  $1 \times 1$ and $3\times 3$     kernels right at the beginning of the network. This  helps us capture the global context alongside with more detailed spatial information. 
 
\textbf{Dilated Convolution}: Receptive fields play a crucial role in extracting spatial features. A  straightforward approach to capture more descriptive spatial features would be to enlarge the kernel size. This has, however, a drawback of increasing the number of parameters drastically. Instead, we replace  the ResNet blocks in the original \sn encoder with a novel combination of a set of dilated convolutions having effective receptive fields of $3, 5$ and $7$ (see Block I in Fig.~\ref{fig:salsanext}). We further concatenate each  dilated convolution output and apply a $1 \times 1$ convolution followed by a residual connection in order to let the network   exploit more information from the fused features coming from various depths in the receptive field. 
Each of these new residual dilated convolution blocks (\ie Block I) is followed by dropout and pooling layers as depicted in Block II in Fig.~\ref{fig:salsanext}. 

\textbf{Pixel-Shuffle Layer}:
The original \sn decoder involves transpose convolutions which are computationally expensive layers in terms of number of parameters. We replace these standard transpose convolutions with the \textit{pixel-shuffle} layer \cite{pixelshuffle} (see Block III in Fig.~\ref{fig:salsanext}) which leverages on the learnt feature maps to produce the upsampled feature maps by shuffling the pixels from the channel dimension to the spatial dimension. More precisely, the \textit{pixel-shuffle} operator reshapes the elements of $(H \times W \times Cr^2)$ feature map to a form of $(Hr \times Wr \times C)$, where $H, W, C,$ and $r$ represent the height, width, channel number and upscaling ratio, respectively. 

We additionally double the filters in the decoder side and concatenate the \textit{pixel-shuffle} outputs with the skip connection (Block IV in Fig.~\ref{fig:salsanext}) before feeding them to the dilated convolutional blocks (Block V in Fig.~\ref{fig:salsanext}) in the decoder.

\textbf{Central Encoder-Decoder Dropout}:
As shown by quantitative  experiments in \cite{kendall2015bayesian}, inserting dropout  only to the  central encoder and decoder layers results in better segmentation performance. It is because the lower network layers extract   basic features such as edges and corners \cite{ZeilerF13} which are consistent over the data distribution and dropping out these layers will prevent the network to properly form the higher level features in the deeper layers. Central dropout approach eventually leads to higher network performance.
We, therefore, insert dropout in every encoder-decoder layer except the first and last one highlighted by dashed edges in Fig.~\ref{fig:salsanext}.


\textbf{Average Pooling}:
In the base \sn model the downsampling was performed via a strided convolution which introduces additional learning parameters. 
Given that the down-sampling process is relatively straightforward, we hypothesize that learning at this level would not be needed. Thus, to allocate less memory \snx switches to average pooling for the downsampling.

All these  contributions from the proposed \snx network. Furthermore,  we applied a $1 \times 1$ convolution   after the decoder unit   to make the channel numbers the same with the total number of semantic classes. The final feature map is finally passed to a soft-max classifier to compute pixel-wise classification scores. 
Note that each convolution layer in the  \snx model employs a leaky-ReLU activation function and is followed by batch normalization to solve the internal covariant shift.    
Dropout is then placed after the batch normalization. It can, otherwise, result in a shift in the weight distribution which can minimize the batch normalization effect during training as  shown in \cite{li2018understanding}. 

\subsection{Uncertainty Estimation}
\label{sec:uncertest}

\subsubsection{Heteroscedastic Aleatoric Uncertainty}
We can define \textit{aleatoric} uncertainty as being of two kinds: \textit{homoscedastic} and \textit{heteroscedastic}. The former defines the type of \textit{aleatoric} uncertainty that remains constant given different input types, whereas the later  may rather differ for different types of input. 
In the LiDAR semantic segmentation task, distant points might introduce a  \textit{heteroscedastic} uncertainty as it is increasingly difficult to assign them to a single class. The same kind of uncertainty is also observable in the object edges when performing semantic segmentation, especially when the gradient between the object and the background is not sharp enough.



LiDAR observations are usually corrupted by noise and thus the input that a neural network is processing is a noisy version of the real world. Assuming that the sensor's noise characteristic is known (\eg  available in the sensor data sheet), the input data distribution can be expressed by the normal $\mathcal{N}(\mathbf{x}, \mathbf{v})$, where $\mathbf{x}$ represents the observations and $\mathbf{v}$ the sensor's noise. In this case, the aleatoric uncertainty can be computed by propagating the noise through the network via Assumed Density Filtering (ADF). This approach was initially applied 
by Gast \textit{et al.}~\cite{gast2018lightweight}, where the network's activation functions including   input and   output were replaced by probability distributions. A forward pass in this ADF-based modified neural network finally generates output predictions $\mathbf{\mu}$  with their respective aleatoric   uncertainties $\mathbf{\sigma}_{A}$.

\subsubsection{Epistemic Uncertainty}
In \snxk the \textit{epistemic} uncertainty is computed using the weight's posterior $p(\mathbf{W|X,Y)}$ which is intractable and thus impossible to present analytically. However, the work in \cite{gal2016dropout} showed that dropout can be used as an approximation to the intractable posterior. More specifically, dropout is an approximating distribution $q_\theta(\omega)$ to the posterior in a BNN with $L$ layers, $\omega = [\mathbf{W}_l]_{l=1}^{L}$ where $\theta$ is a set of variational parameters. The optimization objective function can be written as:
\begin{align*}
    \mathcal{\hat{L}}_{MC}(\theta) = -\frac{1}{M}\sum_{i\in S}\log p(y_i|f^{\omega}(x_i)) + \frac{1}{N}\mathbf{KL}(q_{\theta}||p(\omega))
\end{align*}
where the $\mathbf{KL}$ denotes the regularization from the Kullback-Leibler divergence, $N$ is the number of data samples, $S$ holds a random set of $M$ data samples, $y_i$ denotes the ground-truth, $f^{\omega}(x_i)$ is the output of the network for $x_i$ input with weight parameters $\omega$ and $p(y_i|f^{\omega}(x_i))$ likelihood. 
The $\mathbf{KL}$ term can be approximated as:
\begin{align*}
    KL(q_{M}(\mathbf{W}) || p(\mathbf{W})) \propto \frac{i^2(1 - p)}{2}||\mathbf{M}||^2 - K\mathcal{H}(p)
\end{align*}
 where 
\begin{align*}
    \mathcal{H}(p) := -p\log(p) - (1 -p)\log(1-p)
\end{align*}
represents the entropy of a Bernoulli random variable with probability $p$ and $K$ is a constant to balance the regularization term with the predictive term.

For example, the negative log likelihood in this case will be estimated as 
\begin{align*}
    -\log p(y_i|f^{\omega}(x_i)) \propto \frac{1}{2}\log\sigma  + \frac{1}{2\sigma} ||y_i - f^{\omega}(x_i)||^2 
\end{align*}
for a Gaussian likelihood with $\sigma$ model's uncertainty.




To be able to measure the \textit{epistemic} uncertainty, we employ a Monte Carlo sampling during inference: we run $n$ trials and compute the average of the variance of the $n$ predicted outputs:
\begin{align*}
    \text{Var}^{epistemic}_{p(y|f^{\omega}(x))} = \sigma_{epistemic} =  \frac{1}{n}\sum_{i = 1}^n(y_i - \hat{y})^2~\mathpunkt
\end{align*}

As introduced in~\cite{segu2019general}, the optimal dropout rate $p$ which minimizes the $\mathbf{KL}$ divergence, is estimated for an \textit{already trained network} by applying a grid search on a log-range of a certain number of possible rates in the range $[0, 1]$. In practice, it means that the optimal dropout rates $p$ will minimize:
\begin{align*}
    p = \argmin_{\hat{p}} \sum_{d\in D} \frac{1}{2} \log(\sigma_{tot}^d) + \frac{1}{2\sigma_{tot}^d}(y^d - y_{pred}^d(\hat{p}))^2 ~\mathkomma
\end{align*}

where $\sigma_{tot}$ denotes the total uncertainty by summing the aleatoric and the epistemic uncertainty, $D$ is the input data, $y_{pred}^d(\hat{p})$ and $y^d$  are the predictions and  labels, respectively.

\subsection{Loss Function}
\label{sec:loss}

Datasets with imbalanced classes introduce a challenge for neural networks.
Take an example of  a bicycle  or  traffic sign which appears much less compared to the vehicles in the autonomous driving scenarios. This makes the network more biased towards to the classes that emerge more in the training data and thus yields significantly  poor network performance. 


To cope with the imbalanced class problem, we follow the same strategy in \sn and add more value to the under-represented classes by weighting the softmax cross-entropy loss $\mathcal{L}_{wce}$ with the inverse square root of class frequency  as
\begin{align*}
\resizebox{1.0\hsize}{!}{$\mathcal{L}_{wce}(y,\hat{y}) = -\sum_{i} \alpha_{i}p(y_{i})log(p(\hat{y}_{i}))  ~~~ with ~~~
        \alpha_{i} = 1/\sqrt{f_{i}}    \mathkomma$}
\end{align*}
where $y_i$ and $\hat{y}_{i}$ define the true and predicted class labels and $f_{i}$ stands for the frequency, \ie the number of points, of the $i^{th}$ class. 
This reinforces the network response to the  classes appearing less in the dataset.

In contrast to \snk we here also incorporate the \ls loss \cite{berman2018lovasz}   in the learning procedure to maximize the intersection-over-union (IoU) score, \ie the Jaccard
index.
The IoU metric (see section~\ref{metrics}) is the most commonly used metric to evaluate the segmentation performance. 
Nevertheless, IoU is a discrete and not derivable metric that does not have a direct way to be employed as a loss. In \cite{berman2018lovasz}, the authors adopt this metric with the help of the Lov\'{a}sz extension for submodular functions. 
Considering the IoU  as a hypercube where each vertex is a possible combination of the class labels, we relax the IoU score to be defined everywhere inside of the hypercube. In this respect, 
the \ls loss ($\mathcal{L}_{ls}$) can be formulated as follows:
\begin{equation*}
\scalemath{0.77}{
\mathcal{L}_{ls} = \frac{1}{|C|}\sum_{c\in C} \overline{\Delta_{J_c}}(m(c)) ~,~~ and ~~~
m_i(c) = \left\{
\begin{array}{l l}
  1-x_i(c) &   \text{if ~ $c = y_i(c)$    } \\
  x_i(c) &   \text{otherwise}\\
\end{array}
\right.
~,
}
\end{equation*}
 
where $|C|$ represents the class number, $\overline{\Delta_{J_c}}$ defines the Lov\'{a}sz extension of the Jaccard index, $x_i(c) \in [0,1]$ and $y_i(c) \in \{-1,1\}$ hold the predicted probability and ground truth label of pixel $i$ for class $c$, respectively.

Finally, the total loss function of \snx is a linear combination of both weighted cross-entropy and \ls losses
as follows:
%
$\mathcal{L} =  \mathcal{L}_{wce} + \mathcal{L}_{ls} $.

%
%

\subsection{Optimizer And Regularization}
As an optimizer, we employed stochastic gradient descent with an initial learning rate of $0.01$ which is decayed by $0.01$ after each epoch. 
We also applied an L2 penalty with $\lambda = 0.0001$ and a momentum of $0.9$. 
%
The batch size and spatial dropout probability  were fixed at $24$ and   $0.2$, respectively.
%

To prevent overfitting, we augmented the data by applying a random rotation/translation, flipping randomly around the y-axis and randomly dropping points before creating the projection. Every augmentation is applied independently of each other with a probability of $0.5$.
\subsection{Post-processing}
\label{sec:postproc}

The main drawback of the projection-based point cloud representation  is the information loss   
due to discretization errors and blurry convolutional layer responses. 
This problem emerges when, for instance, the RV image  is re-projected back to the original 3D space. The reason is that during the image rendering process, multiple LiDAR points may get assigned to the very same image pixel which leads to misclassification of, in particular,   the object edges.  This effect becomes more obvious, for instance, when the objects cast a shadow in the background scene. 

To cope with these back-projection related issues, we employ the kNN-based post-processing technique introduced in \cite{rangenetpp}. 
The post-processing is applied to every LIDAR point by using a window around each corresponding image pixel, that will be translated into a subset of point clouds. Next, a set of closest neighbors is selected with the help of kNN. 
The assumption behind using the range instead of the Euclidian distances lies in the fact that a small window is applied, making the range of close $(u,v)$ points serve as a good proxy for the Euclidian distance in the three-dimensional space.
For more details, we refer the readers to \cite{rangenetpp}.

Note that this post-processing is applied to the network output during inference only and has no effect on   learning.

\begin{table*}\centering
\scalebox{0.74}{
\begin{tabular}{@{}r|l|c|ccccccccccccccccccc|c@{}}
\toprule
& Approach & Size 
&\rotatebox{90}{car}		&\rotatebox{90}{bicycle}		&\rotatebox{90}{motorcycle}		&\rotatebox{90}{truck}			&\rotatebox{90}{other-vehicle}
&\rotatebox{90}{person} 	&\rotatebox{90}{bicyclist}		&\rotatebox{90}{motorcyclist} 	&\rotatebox{90}{road}			&\rotatebox{90}{parking} 	
&\rotatebox{90}{sidewalk}   &\rotatebox{90}{other-ground} 	&\rotatebox{90}{building} 		&\rotatebox{90}{fence} 			&\rotatebox{90}{vegetation} 
&\rotatebox{90}{trunk} 		&\rotatebox{90}{terrain} 		&\rotatebox{90}{pole} 			&\rotatebox{90}{traffic-sign}	&\rotatebox{90}{\textbf{mean-IoU}}    
\Bstrut\\ 
\hline   

\Tstrut \multirow{7}{*}{\rotatebox[origin=r]{90}{\textit{Point-wise}}}   & Pointnet~\cite{pointnet} &\multirow{6}{*}{50K pts} & 
46.3 & 1.3 & 0.3 & 0.1 &0.8 & 0.2 & 0.2 & 0.0 & 61.6 & 15.8 & 35.7 & 1.4 & 41.4 & 12.9 & 31.0 & 4.6 & 17.6 & 2.4 & 3.7 & 14.6 \\
& Pointnet++~\cite{pointnetpp} && 
53.7 &1.9 &0.2 &0.9 &0.2& 0.9 &1.0 &0.0 &72.0 &18.7& 41.8 &5.6 &62.3 &16.9 &46.5 &13.8 &30.0& 6.0& 8.9& 20.1 \\
& SPGraph~\cite{SPGraph} &&  
68:3 &0.9 &4.5 &0.9 &0.8& 1.0& 6.0 &0.0 &49.5& 1.7& 24.2& 0.3& 68.2& 22.5 &59.2 &27.2 &17.0 &18.3 &10.5 &20.0\\
& SPLATNet~\cite{SPLATNet} &&  
66.6 & 0.0 &0.0 &0.0 &0.0 &0.0 &0.0 &0.0 &70.4 &0.8 &41.5 &0.0& 68.7 &27.8 &72.3 &35.9 &35.8 &13.8& 0.0& 22.8\\
& TangentConv~\cite{TangentConv} && 
86.8 &1.3 &12.7 &11.6 &10.2 &17.1 &20.2 &0.5 &82.9 &15.2 &61.7 &9.0 &82.8& 44.2 &75.5 &42.5 &55.5 &30.2 &22.2 &35.9\\
& RandLa-Net~\cite{Randlanet} && 
\textbf{94.2} &  26.0 &  25.8 & \textbf{40.1} &\textbf{38.9 }&49.2 & 48.2 &7.2 & 90.7 &60.3 & 73.7 &\textbf{38.9} &86.9 & 56.3 & 81.4 & 61.3 &\textbf{66.8} &49.2 &47.7 & 53.9 \\ 
& LatticeNet~\cite{LatticeNet} &&   
92.9 & 16.6 & 22.2 & 26.6 & 21.4& 35.6 & 43.0& \textbf{46.0} & 90.0 & 59.4 & 74.1 & 22.0 & 88.2 & 58.8 & 81.7 & \textbf{63.6} & 63.1 & 51.9& 48.4&
~52.9 
\Bstrut\\ 
\hline
\Tstrut  \multirow{12}{*}{\rotatebox[origin=r]{90}{\textit{Projection-based}}} & SqueezeSeg~\cite{SqueezesegV01} &\multirow{8}{*}{\shortstack{64$\times$2048\\ pixels}} & 
68.8 &16.0 &4.1 &3.3 &3.6 &12.9 &13.1 &0.9 &85.4 &26.9 &54.3 &4.5 &57.4 &29.0 &60.0 &24.3 &53.7 &17.5 &24.5 &29.5 \\
& SqueezeSeg-CRF~\cite{SqueezesegV01} && 
68.3 &18.1 &5.1 &4.1 &4.8 &16.5 &17.3 &1.2 &84.9 &28.4 &54.7 &4.6 &61.5 &29.2 &59.6 &25.5 &54.7 &11.2 &36.3 &30.8 \\
& SqueezeSegV2~\cite{SqueezesegV02} && 
81.8& 18.5& 17.9 &13.4& 14.0 & 20.1 &25.1 &3.9 &88.6 &45.8 &67.6 &17.7 &73.7 &41.1 &71.8 &35.8& 60.2 &20.2 &36.3 &39.7 \\
& SqueezeSegV2-CRF~\cite{SqueezesegV02} && 
82.7 &21.0 &22.6 &14.5 &15.9 &20.2 &24.3 &2.9 &88.5 &42.4 &65.5 &18.7 &73.8 &41.0 &68.5 &36.9 &58.9 &12.9 &41.0 &39.6 \\
& RangeNet21~\cite{rangenetpp} && 
85.4 &26.2 &26.5 &18.6 &15.6 &31.8 &33.6 &4.0 &91.4 &57.0 &74.0 &26.4 &81.9 &52.3 &77.6 &48.4 &63.6 &36.0 &50.0 &47.4 \\
& RangeNet53~\cite{rangenetpp} && 
86.4 &24.5 &32.7 &25.5 &22.6 &36.2 &33.6 &4.7 & \textbf{91.8} &64.8 &74.6 & 27.9  &84.1 &55.0 &78.3 &50.1 &64.0 &38.9 &52.2 &49.9 \\
& RangeNet53++~\cite{rangenetpp} && 
91.4 &25.7 &34.4 &25.7 &23.0 &38.3 &38.8 &4.8 & \textbf{91.8} & \textbf{65.0} & 75.2 &27.8 &87.4 &58.6 &80.5 &55.1 &64.6 &47.9 &55.9 &52.2 \\
& 3D-MiniNet~\cite{3Dmininet} &&
90.5 & 42.3 &  \textbf{42.1} & 28.5 & 29.4&  47.8 & 44.1 & 14.5 & 91.6 & 64.2 & 74.5 &  25.4 & 89.4 & 60.8 &  \textbf{82.8} &60.8& 66.7& 48.0 &56.6 & 55.8 \\
& SqueezeSegV3~\cite{SqueezesegV03} &&   
92.5 & 38.7 & 36.5 & 29.6 & 33.0 & 45.6 & 46.2 & 20.1 & 91.7 & 63.4  & 74.8 & 26.4 & 89.0 & 59.4 & 82.0 & 58.7 & 65.4 & 49.6 & 58.9 & ~55.9
\Bstrut\\
\cmidrule{2-23}
%
%
& SalsaNet~\cite{salsanet2020} &\multirow{2}{*}{\shortstack{64$\times$2048\\ pixels}} &  
87.5 & 26.2 & 24.6 & 24.0 & 17.5 & 33.2 & 31.1 & 8.4 & 89.7 & 51.7 & 70.7 & 19.7 & 82.8 & 48.0 & 73.0 & 40.0 & 61.7 & 31.3 & 41.9 & 45.4 \\
& SalsaNext  [Ours] && 
91.9 & \textbf{48.3} & 38.6 & 38.9 & 31.9 & \textbf{60.2} & \textbf{59.0} & 19.4 & 91.7 & 63.7 & \textbf{75.8}& 29.1 & \textbf{90.2} & \textbf{64.2}  &81.8 & \textbf{63.6} & 66.5 & \textbf{54.3} & \textbf{62.1}&\textbf{59.5} \\
%
%
\bottomrule
\end{tabular}
}
\caption{Quantitative comparison on \sk test set (sequences 11 to 21). IoU scores are given in percentage ($\%$).} 
\label{tab:quanresults}
\end{table*}
 
\section{Experiments}

We evaluate the performance of \snx and compare with the other state-of-the-art semantic segmentation methods on the large-scale challenging  \sk dataset~\cite{semantickitti} 
which provides over 43K point-wise annotated full 3D LiDAR scans. 
We follow exactly the same protocol in \cite{rangenetpp} and divide the dataset into training, validation, and test splits. Over 21K scans (sequences between 00 and 10) are used for training, where scans from sequence 08 are particularly dedicated to validation. The remaining scans (between sequences 11 and 21) are used as test split. The dataset has in total $22$ classes $19$ of which are evaluated on the test set by the official online benchmark platform.
We implement our model in PyTorch and release the code for public use~\href{https://github.com/TiagoCortinhal/SalsaNext}{https://github.com/TiagoCortinhal/SalsaNext}

\subsection{Evaluation Metric}
\label{metrics}
To evaluate the results of our model we use the Jaccard Index, also known as mean intersection-over-union (IoU) over all classes that is given by 
$
mIoU = \frac{1}{C} \sum_{i=1}^C \frac{|\mathcal{P}_i \cap \mathcal{G}_i|}{| \mathcal{P}_i \cup \mathcal{G}_i|} 
$
, where $\mathcal{P}_i$ is the set of point with a class prediction $i$, $\mathcal{G}_i$ the labelled set for class $i$ and $|•|$ the cardinality of the set. 


\begin{figure}[!b]
    \centering
    \includegraphics[scale=0.32]{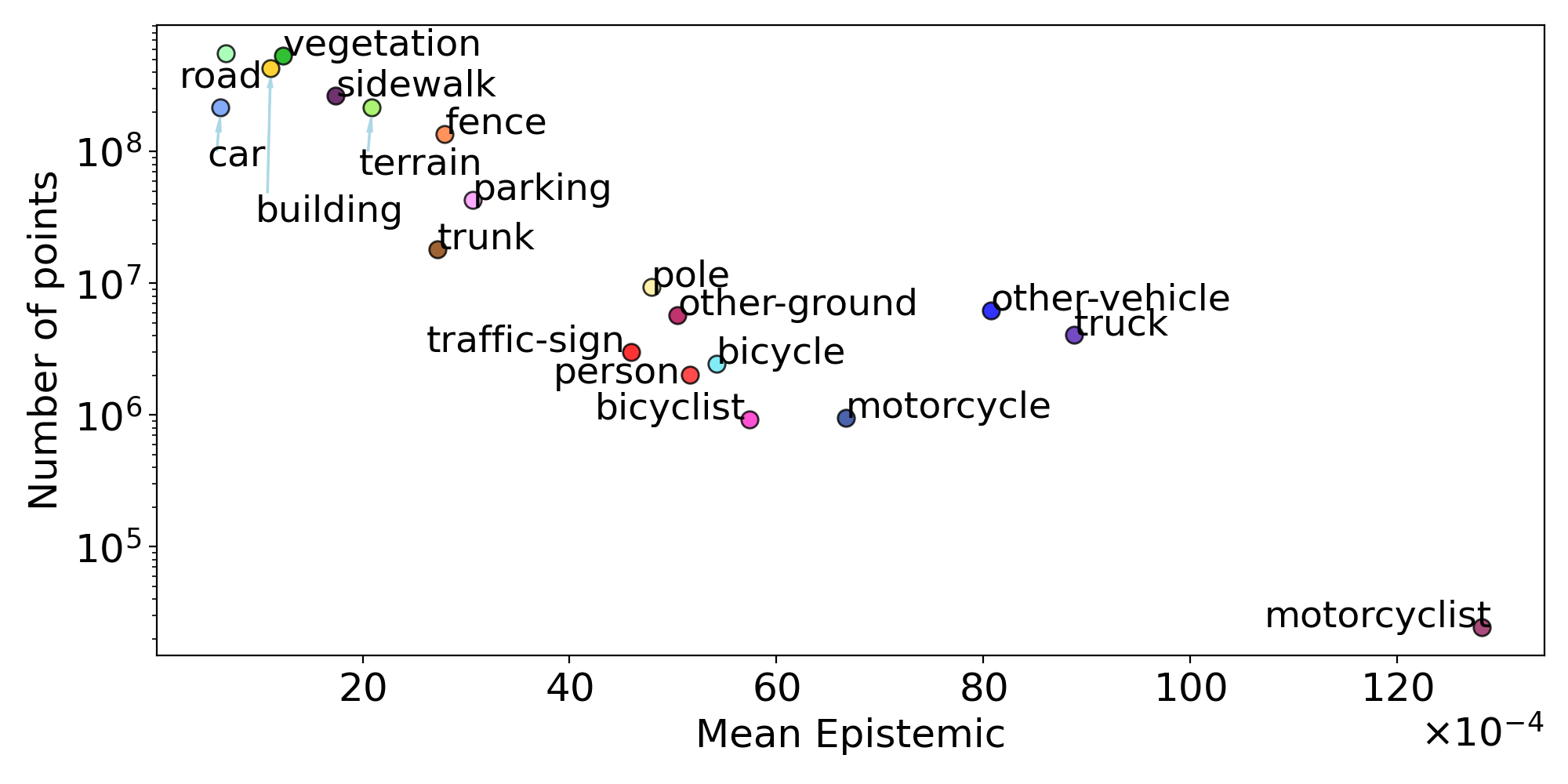}
    \caption{The relationship between the \textit{epistemic} (model) uncertainty and the number of points (in log scale) that each class has in the entire test dataset.  }
    \label{fig:epistuncert}
\end{figure}

\begin{figure*}[!b]
    \centering
    \includegraphics[scale=0.25]{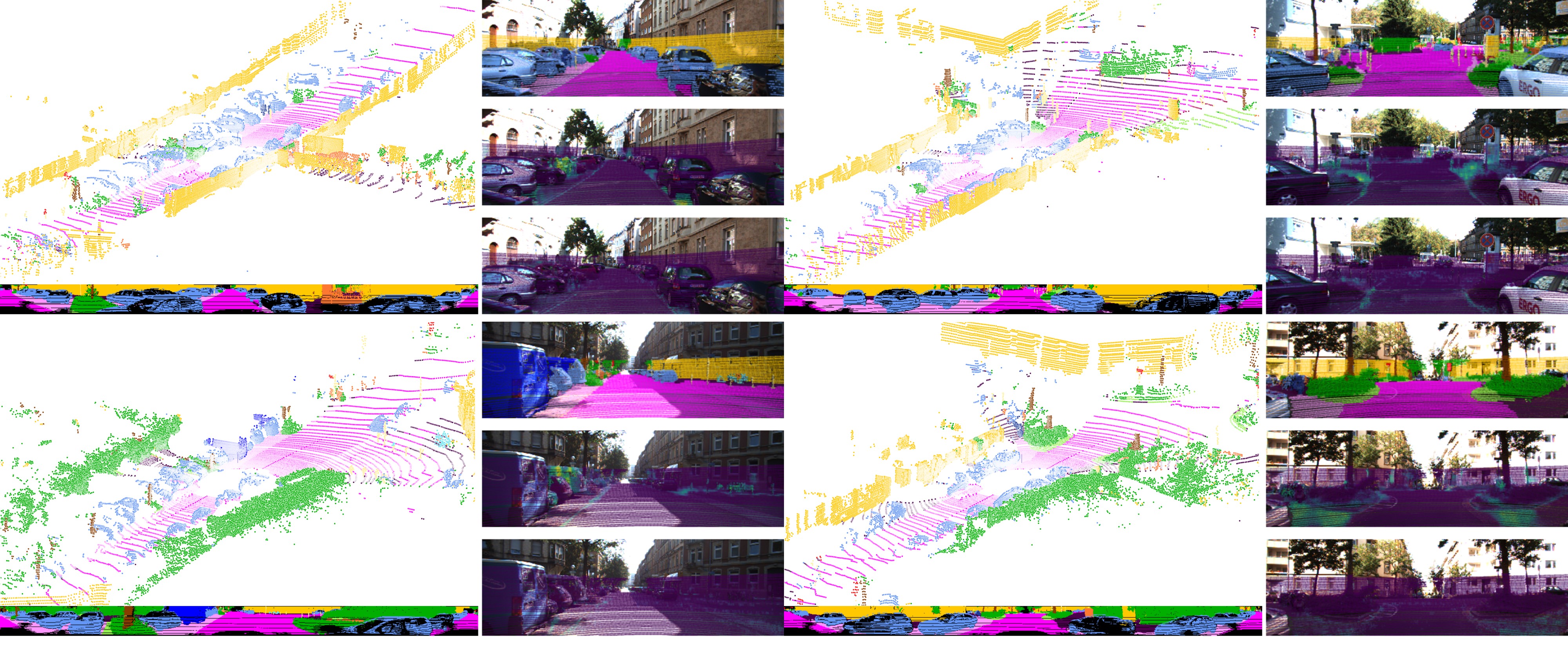}
    \caption{Sample qualitative results showing successes of our proposed \snx method [best view in color]. At the bottom of each scene, the range-view image of the network response is shown. Note that the corresponding camera images on the   right are only for visualization purposes and have not been used in the training. The top camera image on the right shows the projected segments whereas the middle and bottom images depict the projected \textit{epistemic} and \textit{aleatoric}  uncertainties, respectively. Note that the lighter the color is, the more uncertain the network becomes.}
    \label{fig:segresults}
\end{figure*}

\subsection{Quantitative Results}
 
Obtained quantitative results compared to other state-of-the-art point-wise and projection-based approaches are reported in Table~\ref{tab:quanresults}.
Our proposed model \snx considerably outperforms the others by leading to the highest mean IoU score ($59.5\%$) which is  $+3.6\%$  over the previous state-of-the-art method \cite{SqueezesegV03}. In contrast to the original \snk we also obtain more than $14\%$ improvement in the accuracy. 
When it comes to the performance of each individual category, \snx performs the best in $9$  out of 19 categories. Note that in most of these remaining $10$ categories (e.g. road, vegetation, and terrain) \snx  has a comparable performance with the other approaches.


Following the work of \cite{segu2019general}, we further computed the \textit{epistemic} and \textit{aleatoric} uncertainty without retraining the \snx model (see sec.~\ref{sec:uncertest}). 
Fig.~\ref{fig:epistuncert} depicts the quantitative relationship between the \textit{epistemic} (model) uncertainty and the number of points that each class has in the entire \sk test dataset. 
This plot has diagonally distributed samples, which clearly shows that the network becomes less certain about rare classes represented by low number of points (\eg motorcyclist and motorcycle).
There also exists, to some degree, an inverse correlation between the obtained uncertainty and the segmentation accuracy: when the network predicts an incorrect label, the uncertainty    becomes   high as in the case of motorcyclist which has the lowest IoU score ($19.4\%$) in  Table~\ref{tab:quanresults}.

\subsection{Qualitative Results}

For the qualitative evaluation, Fig.~\ref{fig:segresults} shows some sample semantic segmentation and uncertainty results generated by \snx on the \sk test set.

In this figure, only for visualization purposes, segmented object points are also projected back to the respective camera image. We, here, emphasize that these camera images have not been used
for training of \snxp
As depicted in Fig.~\ref{fig:segresults}, \snx can, to a great extent, distinguish road, car, and other object points.
In Fig.~\ref{fig:segresults}, we additionally show the estimated   \textit{epistemic}   and  \textit{aleatoric}   uncertainty values projected on the camera image for the sake of clarity. Here, the light blue points indicate the highest
uncertainty whereas darker points represent more certain predictions.
In line with Fig.~\ref{fig:epistuncert}, we obtain high \textit{epistemic} uncertainty for rare classes such as other-ground as shown in the last frame in  Fig.~\ref{fig:segresults}. We also observe that high level of \textit{aleatoric} uncertainty mainly appears  around segment boundaries (see  the second frame in Fig.~\ref{fig:segresults}) and on distant objects (\eg last frame in Fig.~\ref{fig:segresults}).
In the supplementary video\footnote{\href{https://youtu.be/pf9pSEhE7b0}{https://youtu.be/pf9pSEhE7b0}}, we provide more qualitative results.

\begin{table}[!t]\centering
\scalebox{0.9}{
\begin{tabular}{@{}lcccc@{}}\toprule
                                           & mean IoU  & mean IoU   &Number of     \\  
                                           & (w/o kNN) &  (+kNN)  &  Parameters & FLOPs \\ \midrule
 SalsaNet~\cite{salsanet2020}              & 43.5  & 44.8 & 6.58 M & ~51.60 G   \\
 + context module                          & 44.7  & 46.0 & 6.64 M & ~69.20 G  \\
 + central dropout                         & 44.6  & 46.3 & 6.64 M & ~69.20 G  \\
 + average pooling                         & 47.7  & 49.9 & 5.85 M & ~66.78 G \\
 + dilated convolution                     & 48.2  & 50.4 & 9.25 M & 161.60 G  \\
 + Pixel-Shuffle                      	   & 50.4  & 53.0 & 6.73 M & 125.68 G   \\
 + \ls loss                                & \textbf{56.6}  & \textbf{59.5} & 6.73 M & 125.68 G  \\
\bottomrule
\end{tabular}
}
\caption{Ablative analysis.} 
\label{tab:ablation}
\end{table}


\subsection{Ablation Study}
\label{sec:ablation}

In this ablative analysis, we investigate the individual contribution of each  improvements over the original \sn model.
Table~\ref{tab:ablation}  shows the total number of model parameters and FLOPs (Floating Point Operations)   with the obtained mIoU scores on the \sk test set before and after applying the kNN-based post processing   (see section~\ref{sec:postproc}).
 
As depicted in Table~\ref{tab:ablation}, each of our contributions on \sn has a unique improvement in the accuracy. The post processing step leads to a certain jump (around $2\%$) in the accuracy. The peak in the model parameters is observed when dilated convolution stack is introduced in the encoder, which is vastly reduced after adding the \textit{pixel-shuffle} layers in the decoder. Combining the weighted cross-entropy loss with  \ls leads to the highest increment in the accuracy as the Jaccard index is directly optimized.
We can achieve the highest accuracy score of $59.5\%$ by having only $2.2\%$ (\ie $0.15$M) extra parameters compared to the original \sn model. Table~\ref{tab:ablation} also shows that the number of FLOPs is  correlated with the number of parameters.
%
We note that adding  the \textit{epistemic}  and \textit{aleatoric} uncertainty computations do not introduce any additional training parameter  since they are computed after the network is trained.

\begin{table}[!t]\centering
\scalebox{0.83}{
\begin{tabular}{@{}lcccccc@{}}\toprule
                            & \multicolumn{3}{c}{Processing Time (msec)  } & &    \\
                            \cmidrule{2-4}
                            & CNN  & kNN  & Total     & Speed (fps) & Parameters & FLOPs \\ \midrule
RangeNet++~\cite{rangenetpp}       & ~63.51 & ~2.89 & ~66.41 & 15 Hz & 50 M & 720.96 G\\
SalsaNet~\cite{salsanet2020}       & ~35.78 & ~2.62 & ~38.40 & 26 Hz & 6.58 M & ~51.60 G\\
SalsaNext [Ours]                   & ~38.61 & ~2.65 & ~41.26 & 24 Hz & 6.73 M & 125.68 G\\
\bottomrule
\end{tabular}
}
\caption{Runtime performance on the  Semantic-KITTI test set} 
\label{tab:runtime}
\end{table}
  
\subsection{Runtime Evaluation} 
Runtime performance is of utmost importance in autonomous driving. 
Table~\ref{tab:runtime} reports the total runtime performance   for the CNN backbone network and post-processing module of  \snx in contrast to other networks. To obtain fair statistics, all measurements are performed using the entire \sk dataset on the same single NVIDIA Quadro RTX 6000 - 24GB card. As depicted in  Table~\ref{tab:runtime}, our method clearly exhibits better performance compared to, for instance, RangeNet++~\cite{rangenetpp} while having $7\times$ less parameters.
\snx can run at $24$ Hz when the uncertainty computation is excluded for a fair comparison with deterministic models. 
Note that this high speed we reach is significantly faster than the sampling rate  of mainstream LiDAR sensors which typically work at 10 Hz~\cite{KittiDataset}. 
Fig.~\ref{fig:meanIoUvsRunTime} also compares the overall performance of \snx with the other state-of-the-art  semantic segmentation networks  in terms of runtime, accuracy, and  memory consumption. 

 

\section{Conclusion}
 
We presented a new uncertainty-aware semantic segmentation network, named \snxk that can process the full $360^\circ$ LiDAR scan in real-time.
\snx builds up on the \sn model and can   achieve over $14\%$ more accuracy. In contrast to previous methods, \snx returns  $+3.6\%$ better mIoU score.
Our method differs in that \snx can also estimate both data and model-based uncertainty.
%
  


\bibliographystyle{IEEEtran}
\bibliography{SalsaNext}

\end{document}